# Unravelling the Architecture of Membrane Proteins with Conditional Random Fields

Lior Lukov, Sanjay Chawla, Wei Liu, Brett Church, and Gaurav Pandey

University of Sydney, Australia
{lior.lukov, sanjay.chawla, bret.church}@sydney.edu.au,

University of Technology Sydney, Australia

Wei.Liu@uts.edu.au

University of Minnesota, USA
gaurav@cs.umn.edu

**Abstract**

In this paper we will show that the recently introduced graphical model: Conditional Random Fields (CRF) provides a template to integrate micro-level information about biological entities into a mathematical model to understand their macro-level behaviour. More specifically, we will apply the CRF model to an important classification problem in protein science, namely *the secondary structure prediction of proteins based on the observed primary structure*. A comparison on benchmark data sets against twenty eight other methods shows that not only does the CRF model lead to extremely accurate predictions but the modular nature of the model and the freedom to integrate disparate, overlapping and non-independent sources of information, makes the model an extremely versatile tool to potentially "solve" many other problems in bioinformatics. See the originally compiled PDF of this paper at: https://drive.google.com/file/d/1IYF52Wk8m96KIlrQHUVtEBdm0Kw3M40c

# 1 Introduction

An abundant and important class of proteins are the Integral Membrane Proteins (IMPs) which are found permanently embedded in the cell membrane. IMPs control a broad range of events essential to the proper functioning of cells, tissues and organisms including (i) the regulation of ion and metabolite fluxes, (ii) maintaining appropriate contact between cells, connective tissues, extracellular matrices and (iii) transducing extracellular signals to intracellular signalling pathways [5]. In addition, IMPs include families such as G-protein coupled receptors (GPCRs) that are the most common target of prescription drugs [34]. The task of determining a protein's structure, especially transmembrane segment prediction, has turned out to be difficult for IMPs [30, 22, 13]. The shortage of data on the structures of membrane proteins, compared to water soluble proteins, stems from a lack of success of X-ray crystallography and NMR spectroscopy on these proteins. Structural determination of IMPs can be problematic due to difficulties in obtaining sufficient amounts of sample. A further contributing factor is that IMPs are generally much more likely to become inactive while handling or waiting for crystallization [17].

Traditional transmembrane segment prediction methods are based on empirical observations that membrane regions are often 20 to 30 residues long, with high hydrophobicity around those regions and are connected with short loops containing positively charged residues. The methods vary in the way scoring metrics are assigned to the hydrophobicity property and the prediction algorithm. More recently, mathematical and probabilistic models such as neural networks [1] and hidden Markov models [13] have been applied to the transmembrane prediction problem. The most accurate methods claim to predict more than 90% of all membrane regions, with full helix prediction for all proteins higher than 80% [29].

However, as secondary data about of IMPs and amino acids has grown, an opportunity has arisen to integrate these diverse (but often correlated) sources of information towards improving transmembrane prediction. A new method is required which can seamlessly integrate information from hetrogeneous sources and use it towards transmembrane prediction.

In this paper we show that the recently introduced graphical model, Conditional Random Field (CRF) provides such a framework to harmonize micro-level information about biological entities in order to understand their macro-level behavior. CRFs are related to Hidden Markov Models (HMMs) but are more general and do not place statistical restrictions on the type of information that can be integrated into the model [16]. For example, in HMMs an important assumption is that the observations (i.e., data) are independent of each other given their latent state (or label). This is a severe restriction and limits the type of information that can be integrated together [29].

More formally, the problem we address in this paper can be described as follows. Given a set of membrane protein sequences where each single record in the set contains a pair of sequences: the observation sequence consisting of amino acids represented by $x$, and the label sequence represented by $y$. The label sequence consists of indicators 1 or 0 which denote the presence or absence of the alpha-helix structure at that location. An example set of sequence pairs is shown in Fig 1. The objective of the secondary prediction problem is to predict the label sequence $y$ given the observation sequence $x$.

Since CRFs provide a template for integrating information, our solution was to generate a large set of "indicators" (we call them features) and iteratively test the instatiated model against known standard benchmarks. Many of the features were derived using knowledge in the existing literature. To give one example about information integration, we were able to combine data about the hydrophobicity of amino acids and whether there are net electron donors or acceptors and test whether integrating these two pieces of information improves the predictability of the model. In the end we settled on a model which had an accuracy of 88% and 84% for segment and amino acid prediction respectively

```
MFINRWLFSTNHKDIGTLYLLFGAWAGMVGTALSLLIRAELGQPGTLLGDDQIYNVVVTAHAFV
0000000000011111111111111111111111111111000000000011111111111111

MAYPMQLGFQDATSPIMEELLHFHDHTLMIVFLISSLVLYIISLMLTTKLTHTSTMDAQEVETI
0000000000000000001111111111111111111111110000000000000000011111

MTHQTHAYHMVNPSPWPLTGALSALLMTSGLTMWFHFNSMTLLMIGLTTNMLTMYQWWRDVIRE
0000000000000001111111111111111111000000111111111111111111111111

MENLNMDLLYMAAAVMMGLAAIGAAIGIGILGGKFLEGAARQPDLIPLLRTQFFIVMGLVDAIP
0000000000111111111111111111111000000000000000000000111111111111

MNGTEGPNFYVPFSNKTGVVRSPFEAPQYYLAEPWQFSMLAAYMFLLIMLGFPINFLTLYVTVQ
0000000000000000000000000000000000001111111111111111111111111110
```

Figure 1: Examples of membrane proteins sequence pairs. The top of the pair consists of a sequence of amino acids and the bottom of the pair is labeled 1 or 0 depending upon whether the amino acid adopts the alpha-helix structure or not at that location. The objective is to predict the label sequence on the basis of the observation sequence.

on a standard benchmark for TMH prediction. In comparsion with twenty eight other methods hosted at the benchmark website[12], the CRF model had the highest accuracy.

While the CRF model demonstrates high overall accuracy, we also wanted to test how it performs on a particular protein structure. Towards that we have thoroughly analyzed the a well known protein complex known as Cytochrome c oxidase. This is a large and well studied protein complex consisting of twenty eight transmembrane regions [28]. Our experiments on this complex using the CRF model predicted the distribution of isoleucene to be similar to that observed in the experimental analysis of Wallin et al [33]. This demonstrates that the CRF model is appropriate for addressing the problem of predicting the structure of IMPs and similar biological problems.

# 2 The CRF Mathematical Model

The sequential classification problem appears in many different fields including computational linguistics [18], speech recognition [23], and computational biology. The underlying structure is as follows: given an observation sequence, the objective is to find (or learn) a corresponding label sequence. For example, in computational linguistics, the observation sequence is the sequence of words and the label sequence are the Part-of-Speech (POS) tags. In speech recognition, the observations are acoustic parameters and the label sequence are words. The label sequence is the classification of the items in the observation sequence. As noted before proteins consist of a sequence of amino acids and that they may have a different secondary structure along their polypeptide chain, which can be either $\alpha$-helix, $\beta$-sheets or coils. For IMPs it is the location of the $\alpha$-helix that is considered most important reducing the problem into a binary sequential classification problem [7].

In order to assign labels to observations we need to build a statistical model which relates the two. Let $x$ and $y$ be observation and label sequence respectively. Our objective is to estimate the probability distribution $P(x, y)$. Without making any further assumptions, this estimation problem is intractable. For example, if the size of the observation alphabet is twenty and that of the label alphabet is two and all sequences are of fixed length $n$, then we need to estimate $20^n \times 2^n$ terms.

If the general problem is intractable, the question is what simplifying assumptions can be made on $P(x, y)$ in order to make the problem tractable without loosing the essence of the relationship between $x$ and $y$?

At this point, it is best to relate $P(x, y)$ with an underlying graphical model. For the sequential classification problem, the graphical model is a pair of chains corresponding to the observation and the label sequence. The graphical model is used to encode and represent the factorization of $P(x, y)$. The nodes of the graph correspond to the random variables $x$ and $y$, and two nodes are connected with an edge if the random variables are related or dependent. For example, suppose a sequence is of length $n$, then $x = x_1 x_2 \ldots x_n$ is a sequence of random variables corresponding the amino acids. Similarly $y = y_1 y_2 \ldots y_n$ corresponds to the sequence of secondary structure labels. If the $x_i$'s and $y_j$'s are related then that relationship is encoded in the underlying graph as an edge between the two nodes. Edges may be directed or undirected. Directed edges specify the direction of dependency between the two random variables.

For example in Fig 2(a), the factorization (known as Hidden Markov Model) is carried out as follows (See the originally compiled PDF of this paper at: https://drive.google.com/file/d/1IYF52Wk8m96KIlrQHUVtEBdm0Kw3M40c):

$$p(x, y) = p(y_1)p(x_1|y_1)\prod_{i=2}^{n} p(y_i|y_{i-1})p(x_i|y_i) \quad (1)$$

Similarly in Fig 2(b), a different factorization (known as Maximum Entropy Markov Model(MEMM)[19]) for $p(y|x)$ is encoded as:

$$p(y|x) = p(y_1|x_1)\prod_{i=2}^{n} p(y_i|y_{i-1}, x_i) \quad (2)$$

The pros and cons between the two factorizations has been a subject of intense debate in the data mining and machine learning literature and we refer the interested reader to some recent expositions and references therein [2, 26, 21]. However, we summarize the salient points:

1. Hidden Markov Models (HMMs) are *generative* while MEM models are *discriminative*. This means that HMMs have to make an assumption about how the observation is generated from the labels $P(x_i|y_i)$, while the MEM directly models the classification task and make an assumption on the functional form of $P(y_i|y_{i-1}, x_i)$ which is derived using the maximum entropy principle. The MEM functional form leads to an exponential model

$$p(y|x) = \frac{1}{Z(x, y)} exp\left(\sum_j \lambda_j F_j(y, x)\right) \quad (3)$$

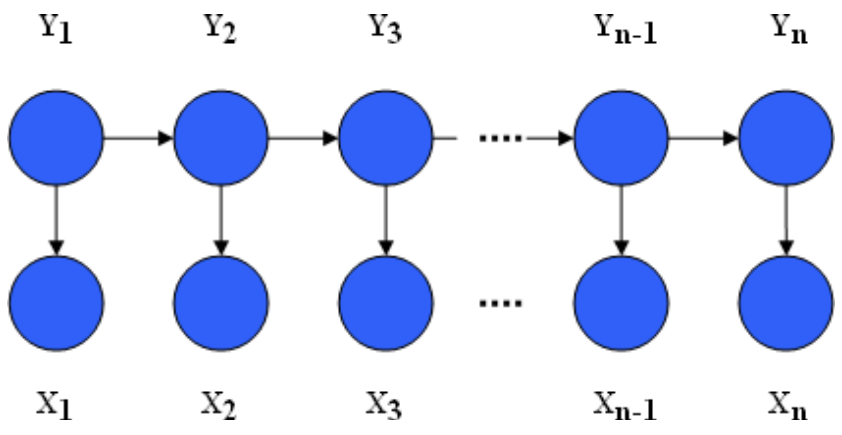

(a) HMM

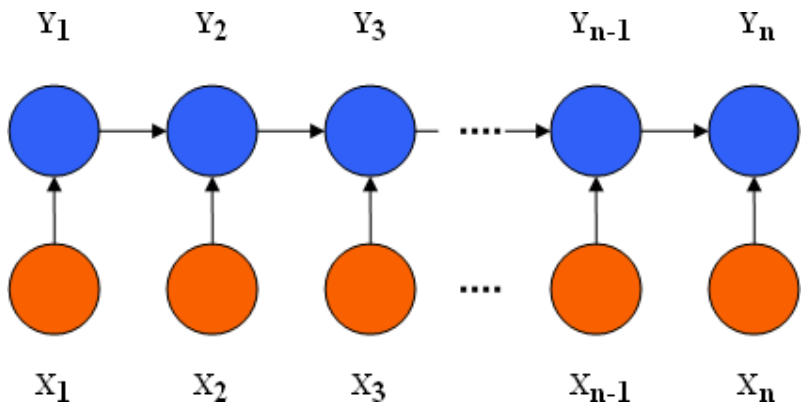

(b) MEMM

Figure 2: Factorization of P(x,y) is necessary in order to make the sequential classification problem tractable. Two common factorizations are the Hidden Markov Model (HMM) and the Maximum Entropy Markov Model (MEMM).

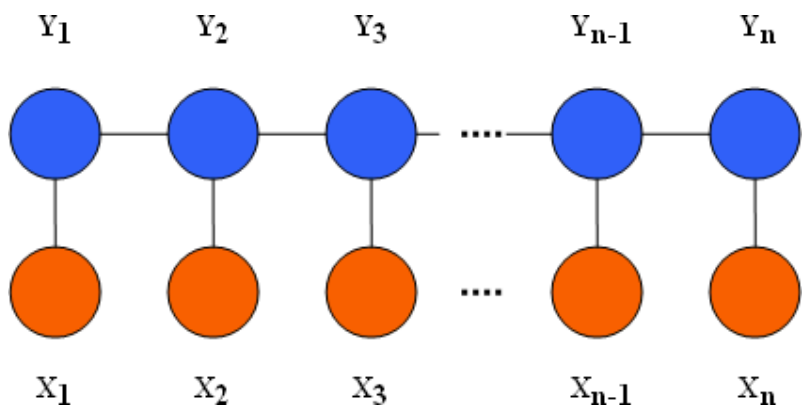

Figure 3: The CRF is an undirected graphical model. The $Y_i$ are conditionally independent given their neighbors. Even though in the figure only $X_i$ is connected to $Y_i$, the CRF model places no restrictions on the interactions between the Y and the X random variables.

1. The term $Z(x, y)$ is known as the normalization term (so that the probabilities add up to one) and because of the fact that it is dependent of both $x$ and $y$ it is called per-state normalization. The depedence of $Z(x, y)$ on both $x$ and $y$ leads to what is called *label-bias*[16]. The CRF model that we will derive will be very similar to the MEM model except that the underlying graphical model will be an undirected chain and the normalizing term will be of the form $Z(x)$. (See the originally compiled PDF of this paper at: https://drive.google.com/file/d/1IYF52Wk8m96KIlrQHUVtEBdm0Kw3M40c)

$$p(y|x) = \frac{1}{Z(x)} exp \Big( \sum_j \lambda_j F_j(y, x) \Big) \quad (4)$$

The label-bias problem occurs, for example, when $P(y_i | y_{i-1}, X) \approx P(y_i | y_{i-1})$. In this case, the MEM model completely ignores the information about the presence or absence of particular amino acids. However, the CRF model takes a more global view and does take the presence of amino acid into account.

2. Both HMMs and the MEM model assume that the label sequence is a markov chain, i.e., the state of the label at $i$, $y_i$ is independent of all previous states except $y_{i-1}$.

3. For the observation sequence, HMMs assume that each observation is independent of all other observations given the labels. The MEM model does not make this assumption. In fact in the MEM model, the relationships between the observations and labels is mediated through the feature functions $F_j(x, y)$ and there are no limitations on the choice of the $F_j$'s. This flexibility allows micro-level biological information to be encoded into the MEM model without having to worry about observation independence. The CRF inherit this flexibility from the MEM model.

# 3 Feature Integration with the Model

The CRF model is a template which needs to be populated with problem dependent features. Features in our context represent function on the combination of the primary sequence and labels. It is helpful to think of features as micro-level information about membrane proteins. The role of the CRF model is to aggregate micro-level features in order to predict macro-level behaviour: namely, the secondary structure of the whole protein. Ideally features should be provided by domain-experts who are familiar with the micro-level behaviour of the entities under examination.

The number of different features which can be applied to a model is infinite. Assembling features which are relevant for the problem is an empirical process which involves many experiments. On each experiment, a combination of features is selected and then the model is evaluated based on its prediction. It is possible to evaluate and score the prediction based on the selected feature combination.

In this Section we present the different features which we have used in our experiments. To reiterate, features always appear as a combination of a particular label (state) and information about the observation sequence. We have selected a set of features to capture biological constraints and divided them into eighteen different groups

Table 1: Amino Acids Hydrophobic Index

| Amino Acid | Hydrophobic Value | Amino Acid | Hydrophobic Value |
|---|---|---|---|
| Lysine (K) | -3.9 | Proline (P) | -1.6 |
| Arginine (R) | -4.5 | Glycine (G) | -0.4 |
| Histidine (H) | -3.2 | Alanine (A) | 1.8 |
| Glutamic(E) | -3.5 | Methionine (M) | 1.9 |
| Glutamine (Q) | -3.5 | Cysteine (C) | 2.5 |
| Aspartic acid (D) | -3.5 | Phenylalanine (A) | 2.8 |
| Asparagine (N) | -3.5 | Leucine (L) | 3.8 |
| Trptophan (W) | -0.9 | Valine (V) | 4.2 |
| Tyrosine (Y) | -1.3 | Isoleucine (I) | 4.5 |
| Serine (S) | -0.8 | | |
| Threonine (T) | -0.7 | | |

## 3.1 Features Extracted from Proteins Primary Structure

### 3.1.1 Start, End and Edge Features

By using these features we capture the probability of starting/ending a sequence with a given label or the transition probability for moving between the four combination of labels (H-H, H-NH, NH-H, NH-NH).

### 3.1.2 Basic Amino Acid Features

This feature captures the probability of amino acids to appear inside helical membranes. Twenty different unigram features exist of this type.

### 3.1.3 Hydrophobic Window Features

One of the most important physico-chemical characteristics of amino acids is hydrophobicity [31].

Hydrophobicity analysis has been shown to be an efficient way to detect transmembrane helices. Based on the work of Kyte and Doolittle [15], each amino acid is assigned a unique hydrophobic value. The values of the hydrophobic index are shown in Table 1.

For this feature a sliding window, consisting of 19 residues, is used to test if the average hydrophobic value of the 19 amino acids inside the window is greater than a certain threshold. 19 has been suggested to be the best number of amino acids in a sliding window [15]. In order to find the most effective threshold, 400 helix-segments and 400 loop-segments are randomly selected from the training dataset (The training dataset is introduced in section 4.1). We calculate and compare the average hydrophobic values for these two segments. As shown in Fig 4, most of the average hydrophobic values of helix-segments are greater than 1.0, while those of loop-segments are mostly lower than 1.0. Hence we use 1.0 as the threshold for this feature.

### 3.1.4 Hydrophilic Window Features

Similar to the Hydrophobic Window Features, the Hydrophilic Window Feature is used for detecting loop segments.

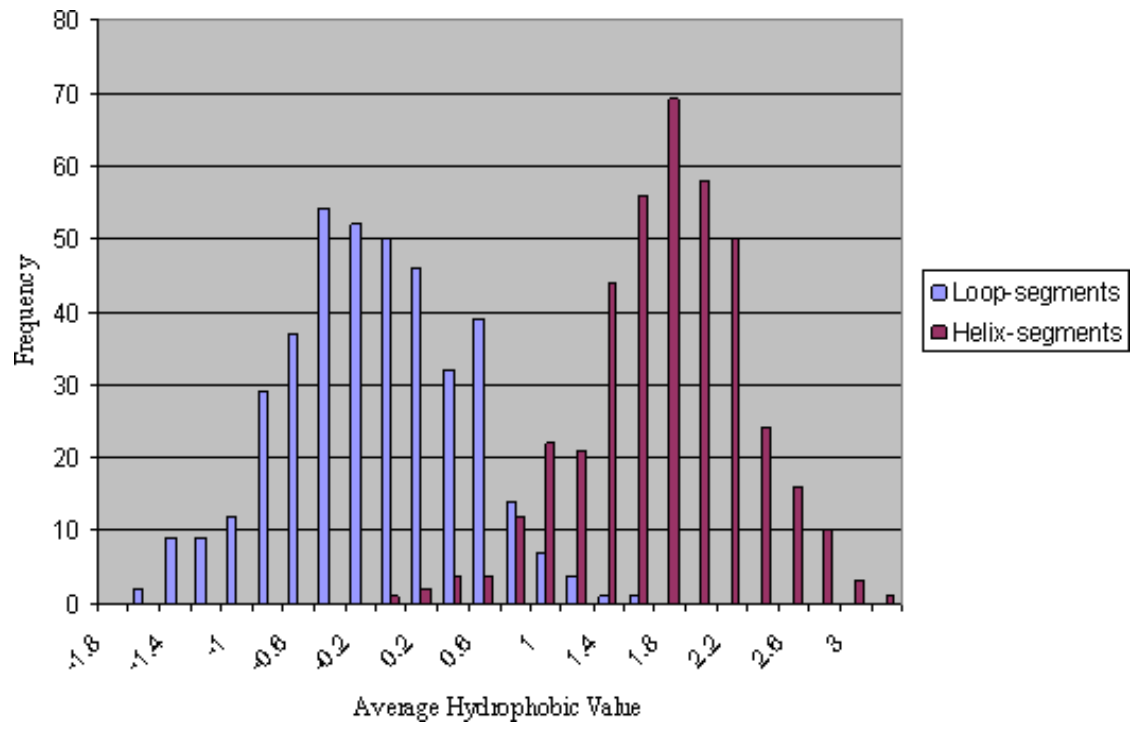

Figure 4: Comparison of Average Hydrophobic Values between Helix Segments and Nonhelix.

#### 3.1.5 Single Side Neighboring Amino Acid Features

While the basic amino acid feature captures the tendency of a given amino acid to appear in a membrane helical structure, we are also interested in capturing the tendency of sequential combinations of amino acids to appear on the left or right of a particular label.

#### 3.1.6 Single Side Shuffled Neighboring Amino Acid Features

Similar to Single Side Neighboring Amino Acid Features, but this time we are interested in capturing the tendency of the same amino acid given its adjacent neighbors without being concerned about their order. The motivation of creating the shuffled features is based on the hypothesis that the location of the transmembrane regions are determined by the difference in the amino acid distribution in various structural parts of the protein rather than by specific amino acid composition of these parts. We test this hypothesis in Section 4.3.2 by comparing the use of Single Side Neighboring Amino Acid Features vs. Single Side Shuffled Neighboring Amino Acid Features on a set of benchmark membrane helix sequences.

#### 3.1.7 Single Side Hydrophobic Neighbouring Amino Acid Features

Similar to Single Side Shuffled Neighbouring Amino Acid Features, but this time we want to capture not only the composition of the adjacent amino acids, but also the hydrophobic tendency of these neighbouring residues. Since the hydrophobic value of a certain segment may affect the formation of a transmembrane helix [31], the hydrophobicity of neighbouring residues may also determine a single amino acid to be in a helix or not. We use the same "threshold" (1.0) as in Hydrophobic Values Window Features, so the Single Side Hydrophobic Neighbouring Amino Acid Feature is only active when the average hydrophobic value of certain neighbouring residues is higher than *1.0*.

### 3.1.8 Single Side Hydrophilic Neighbouring Amino Acid Features

Similar to the Hydrophobic Neighbouring Amino Acid Features, but this time we care about the neighbouring hydrophilic amino acids present in loop-segments. The same "threshold" is used here again, and the Single Side Hydrophilic Neighbouring Amino Acid Feature is only active when the average hydrophilic value of a certain number of neighbouring residues is lower than 1.0.

### 3.1.9 Double Side Neighboring Amino Acid Features

We have also captured the tendency of an amino acid to appear given its adjacent neighbors from both sides together.

### 3.1.10 Double Side Shuffled Neighboring Amino Acid Features

Similar to Double Side Neighboring Amino Acid Features, but this time we are interested in capturing the tendency of an amino acid to appear given adjacent neighbors from both sides together without being concerned about their order in which they appear.

### 3.1.11 Double Side Hydrophobic Neighbouring Amino Acid Features

Similar to Double Side Shuffled Neighbouring Amino Acid Features, but this time we want to capture not only the composition of the adjacent amino acids, but also the hydrophobic tendency of these neighbouring residues from both sides.

### 3.1.12 Double Side Hydrophilic Neighbouring Amino Acid Features

Similar to Double Side Hydrophobic Neighbouring Amino Acid Features, but this time we care about the hydrophilic amino acids present in **loop-segments**.

### 3.1.13 Amino Acid Property Features

Amino acids differ from one another in their chemical structure expressed by their side chains. The fact that amino acids from the same classification group appear in similar locations, motivated us to create special property features. We have adopted the classification from Sternberg [25], classifying the amino acids into nine groups[1], each group described by a unigram feature. Note that some amino acids may appear in more than one group simultaneously.

[1] Aromatic (F,W,Y,H), Hydrophobic (M,I,L,V,A,G,F,W,Y,H,K,C), Positive (H,K,R), Polar (W,Y,C,H,K,R,E,D,S,Q,N,T), Charged (H,K,R,E,D), Negative (E,D), Aliphatic (I,L,V), Small (V,A,G,C,P,S,D,T,N), Tiny (A,G,S)

Table 2: Amino Acid Electronic Property

| *ElectronicProperty* | *AminoAcids* |
|---|---|
| Strong donor | A, D, E, P |
| Weak donor | I, L, V |
| Neutral | C, G, H, S, W, M |
| Weak acceptor | F, Q, T, Y |
| Strong acceptor | K, N, R |

The *Amino Acid Property Feature* highlights a very important fact: This feature and the basic unigram features are clearly dependent. For example, whenever the basic amino acid unigram feature corresponding to the amino acid F is active, the aromatic property feature will be active too (since F is classified as aromatic). Both these features are important. A Hidden Markov Model (HMM) cannot simultaneously handle both these features as it makes a strict assumption about independence between the observations (given the labels).

#### 3.1.14 Border Features

By using these features we capture the border between a segment of amino acids labeled with one structure (helices/loops) and a segment labeled with another (loops/helices).

#### 3.1.15 Short Loop Features

This feature aims at capturing the composition of short loops, which appear to be very difficult to predict especially when they are shorter than 7 residues [4]. In this feature we are not concerned about the order in which the amino acids appear. Instead, we capture the amino acids composition in the short loops. This feature is only active when the Amino Acid in a protein sequence is from a loop segment shorter than 7 residues.

#### 3.1.16 Electron Transport Chain Features

Electron carriers and biochemical reactions are associated by electron transport chains, in which amino acids in protein sequences also take part [8]. The twenty amino acids are classified into five groups, as shown in Table 2, and are used in this feature to capture the influence of electron transport in helix formation.

#### 3.1.17 Chemical Groups Features

Amino acids are characterized by their side chains. Based on the chemical compositions of the side chains, 18 new groups are defined, which form 36 new features, as shown in Table 3.

Table 3: Classification of Amino Acids by the Chemical Component in Their Side Chains

| *GroupNumber* | *ChemicalComponents* | *AminoAcids* |
|---|---|---|
| 1 | $-{-}C-$ | R |
| 2 | $=C^{aromatic}--$ | Y, F, H, W |
| 3 | $--CH-$ | L, V, I, T |
| 4 | $-CH_2-$ | K, N, D, E, L, C, W, S, I, R, Q, F, H, Y |
| 5 | $-CH_2^{ring}-$ | P |
| 6 | $-CH_3$ | L, V, I, A, T, M |
| 7 | $=CH^{aromatic}-$ | W, F, Y, H |
| 8 | $--CH^{ring}$ | W, P |
| 9 | $--C=O$ | N, Q |
| 10 | $-COO-$ | D, E |
| 11 | $=N-$ | H |
| 12 | $-NH-$ | R |
| 13 | $-NH_2$ | N, R, Q |
| 14 | $=NH_2+$ | R |
| 15 | $-NH_3+$ | K |
| 16 | $-OH$ | S, T, Y |
| 17 | $-SH$ | C |
| 18 | $-NH^{ring}-$ | P, H, W |

### 3.1.18 Sequence States Features

The sequence states mechanism in TMHMM [24, 13] have been adopted to be used as features. We divide protein sequences into three states: Helix Core (8 residues in the centre of a transmembrane helix segment), Helix Ends (5 residues at the ends of a transmembrane helix segment), Loops Ends (5 residues at the ends of a loop segment). This feature is based on the hypothesis that the differences between the amino acids distributions in the various structural parts are one of the driving forces in the formation of the transmembrane helices.

## 3.2 CRF Model Prediction Example

We now give a simple example of transmembrane helix prediction. For this example we will assume, for simplicity, that all proteins are 4 amino acids long, and the amino acids alphabet is composed of the amino acids $(A, C, D, E, F) \in X$. The labelling alphabet describes if the current amino acid is inside a helix membrane/non-helix membrane region and denoted by $(0, 1) \in Y$.

We define two property groups:

$$(A,\ C,\ F) \in Hydrophobic$$

$$(C,\ D,\ E) \in Polar$$

Note: the amino acid $C$ is both Hydrophobic and Polar. We create the following unigram features:

$$u_n(x,i) = \begin{cases} 1 & \text{if the acid in sequence } x \text{ at position } i \text{ is from type } n \\ 0 & \text{otherwise} \end{cases}$$

Table 4: Training Data in the Example

| Observation | Labels |
|---|---|
| CAAF | 0111 |
| CDED | 1000 |
| DFAE | 0110 |

Table 5: Activated features on a training set example

| Feature Name | Occurences | Active on sequence |
|---|---|---|
| Basic Helix A | 3 | CAAF, DFAE |
| Basic Helix C | 1 | CDED |
| Basic Non-Helix C | 1 | CAAF |
| Basic Non-Helix D | 3 | CDED, DFAE |
| Basic Non-Helix E | 2 | CDED, DFAE |
| Basic Helix F | 2 | CAAF, DFAE |
| Hydrophobic Helix | 6 | CAAF, CDED, DFAE |
| Hydrophobic Non-Helix | 1 | CAAF |
| Polar Helix | 1 | CDED |
| Polar Non-Helix | 6 | CAAF, CDED, DFAE |

$$u_{Hphobic}(x,i) = \begin{cases} 1 & \text{if the acid at position } i \in (\text{A,C,F}) \\ 0 & \text{otherwise} \end{cases}$$

$$u_{Polar}(x,i) = \begin{cases} 1 & \text{if the acid at position } i \in (\text{C,D,E}) \\ 0 & \text{otherwise} \end{cases}$$

Using these unigrams, each feature for describing the relationship between the observation and the two possible structures has the form:

$$f_{n_H}(y_i, x, i) = \begin{cases} u_n(x,i) & \text{if } y_i = \text{Helix membrane} \\ 0 & \text{otherwise} \end{cases}$$

$$f_{n_{NH}}(y_i, x, i) = \begin{cases} u_n(x,i) & \text{if } y_i = \text{Non-Helix membrane} \\ 0 & \text{otherwise} \end{cases}$$

We describe the relationship among the observation and the two possible structures, helix/non-helix membrane as a feature (as described in the previous Section).

We train the CRFs model using the training set as shown in Table **??**.

Our goal is to predict the labels of the sequence *EAFD*. Table 5 shows the full list of activated features on the given training set.

In the training set a total of ten features are activated. Using maximum likelihood, the parameters $\lambda = (\lambda_1, \lambda_2 \ldots \lambda_{10})$ will be estimted. Table 6 shows the values of these feature parameters after they were calculated.

After calculating the feature parameters, the Viterbi algorithm is applied for labelling the test sequences. Now we will label the sequence *EAFD*. The first letter in the sequence, *E*, is a polar residue. From Table 6 we can calculate the total score of *E* by assigning it with a helical or a non-helical label. For assigning a helical label, the total score of *E* is *Polar Helix* $\approx 0$, while for assigning a non-helical label, the total score is

Table 6: Trained feature parameters

| Feature Name | Parameter value |
|---|---|
| Basic Helix A | 1.7859 |
| Basic Helix C | 5.5170 |
| Basic Non-Helix C | -5.5170 |
| Basic Non-Helix D | 1.7859 |
| Basic Non-Helix E | 1.5391 |
| Basic Helix F | 1.5391 |
| Hydrophobic Helix | 3.3251 |
| Hydrophobic Non-Helix | -5.5170E-8 |
| Polar Helix | 5.5170E-8 |
| Polar Non-Helix | 3.3251 |

Table 7: Calculating the labels of the sequence *EAFD*

| Letter | Label 0 score | Label 1 score | Final score | Assigned Label |
|---|---|---|---|---|
| *E* | 4.8642 | 0 | 4.8642 | 0 |
| *A* | 4.8642 | 9.9752 | 9.9752 | 1 |
| *F* | 9.9752 | 14.8394 | 14.8394 | 1 |
| *D* | 19.9504 | 14.8394 | 19.9504 | 0 |

*Polar Non-Helix + Basic Non-Helix E = 4.8642*. Similarly, the total score of each one of the letters assembling the sequence *EAFD* is calculated for each label. Table 7 summarizes the steps of calculating the sequence labels.

Finally the algorithm assigns for each step these labels which yield the highest score. In this example *EAFD* is assigned with the predicted label sequence of *0110*.

# 4 Experiments, Evaluation and Analysis

In this section we report and analyze the results of three sets of experiments to test the CRF model for transmembrane prediction.

1. In the first experiment we compare different feature selection strategies and their effect on prediction accuracy. The CRF model with the best set of features was then evaluated against twenty eight models hosted at the "Static Benchmarking Server"(henceforth referred as the SBS ) hosted at Columbia University [12].

2. In the second experiment we investigate the hypothesis that the location of the transmembrane region is determined by the distribution of amino acids in the region rather than their ordering. For example, if *FCD* is an observation sub-sequence then the probability that *FCD* is labelled as a transmembrane helix region is similar to the probability that the *permutation* of *FCD* is labelled as a transmembrane helix. This experiment also highlights the modular nature of feature integration into the CRF model. Biological hypothesis and information can be easily translated into features, integrated into the CRF model and tested.

3. The third experiment applies the CRF model on a large and well known

IMP complex, the Cytochrome C Oxidase. We compare our results against those reported, using "wet lab" experiments, by Wallin et. al. [33] and show that we are able to replicate many findings using the CRF-based data driven approach.

Before we describe the experiments, we briefly overview the data sets and the evaluation metrics used throughout this section.

## 4.1 Data Set

The CRF model was trained on a data set consisting of a set of benchmark sequences with experimentally confirmed transmembrane regions compiled by Möller et al. [20]. [2] We have included only proteins with a high level of trust (assigned with transmembrane annotation trust level of A to C, as was suggested by Möller et al.). The resulting set consists of 148 transmembrane protein sequences with both helix and non-helix segments.

The model was tested on SBS [12]. This is a two-step process. First, a data set consisting of 2246 observation sequences was downloaded and labeled by the CRF model. Second, the labeled sequences were uploaded to the website which reported a comparative performance analysis with twenty eight other models. The experimental work flow is illustrated in Fig 5.

## 4.2 Prediction Metrics

The SBS uses two sets of metrics to compare models: per-residue and per-segment accuracy, which we now describe [5, 3].

### 4.2.1 Per-Residue Accuracy

In per-residue accuracy the predicted and actual labels are compared by residue. Let $\Omega_i$ be the sequence of residues in protein $i = 1, \ldots, N_{prot}$. Furthermore, let $\omega_{(i,j)} \in \Omega_i$ be the residue in location $j$ in sequence $\Omega_i$. For simplicity, we denote $\omega_{(i,j)}$ as $\omega$. Let [3]

$$y(\omega) = \begin{cases} 1 & \text{if } \omega \text{ is a TMH residue} \\ 0 & \text{if } \omega \text{ is a NTMH residue} \end{cases}$$

Similarly, let

$$\tilde{y}(\omega) = \begin{cases} 1 & \text{if } \omega \text{ is predicted as a TMH residue} \\ 0 & \text{if } \omega \text{ is predicted as a NTMH residue} \end{cases}$$

Table 8 lists the metrics which capture per-residue accuracy. The symbols in the last column are from Chen et al.[3], and are also used by SBS to report results.

[2] The data set can be accessed via ftp://ftp.ebi.ac.uk/databases/testsets/transmembrane

[3] TMH = Transmembrane Helix, NTMH = Non-Transmembrane Helix

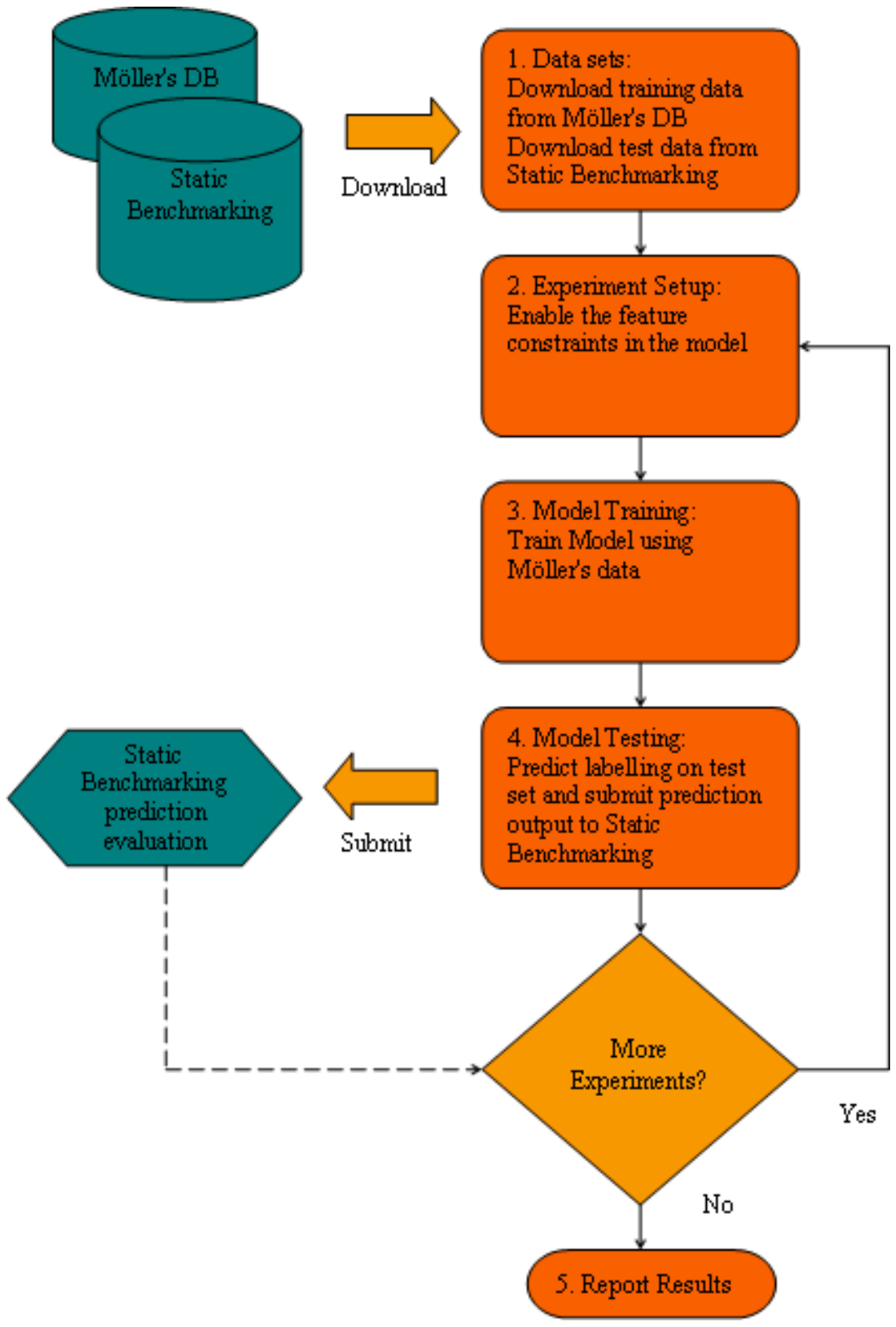

Figure 5: Transmembrane helix prediction experimental flow.

#### 4.2.2 Per-Segment Accuracy

The per-residue accuracy measures ignore the sequential contiguity of the transmembrane helical (TMH) regions. We also want to determine how accurately a method correctly predicts the location of a TMH region.

In order to predict the sequential contiguity of the TMH region we have used the per-segment accuracy metric suggested by Chen et al. [3]. It requires a minimal overlap of **three residues** between the two corresponding segments and does not allow the same helix to be counted twice. For example consider the observed data and two possible prediction sequences: (0 = Non-Transmembrane Helix, 1 = Transmembrane Helix) as shown in Fig 6.

By using the per-segment metric, Prediction1 returns an accuracy of 100% (as it predicts two helices which are assigned with the two observation helices), while Prediction2 returns an accuracy of 50% (as it predicts one helix which is assigned only with the first observation helix). Table 8 also lists the metrics which capture per-segment accuracy.

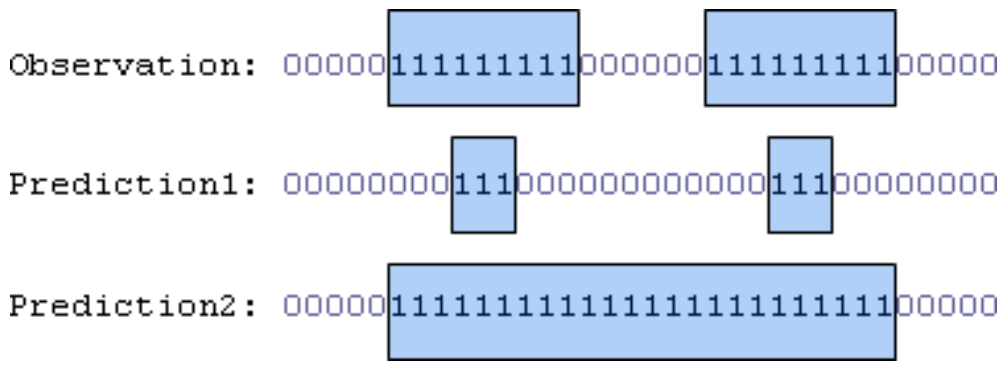

Figure 6: Per-segment accuracy example. Prediction1 is more accurate than Prediction 2 even though Prediction2 has a higher per-residue precision and recall.

Table 8: The metrics used to measure per-residue and per-segment accuracy. $N_{prot}$ = Number of Proteins in the Data Set, TMH = Transmembrane Helix, NTMH = Non-Transmembrane Helix. † These symbols are from Chen et. al.

| Per-Residue Metrics | | |
|---|---|---|
| **Description** | **Formula** | **Symbol†** |
| TMH Recall | $P(\tilde{y}(\omega) = 1 \mid y(\omega) = 1)$ | $Q_{2T}^{\%obs}$ |
| TMH Precision | $P(y(\omega) = 1 \mid \tilde{y}(\omega) = 1)$ | $Q_{2T}^{\%prd}$ |
| NTMH Recall | $P(\tilde{y}(\omega) = 0 \mid y(\omega) = 0)$ | $Q_{2N}^{\%obs}$ |
| NTMH Precision | $P(y(\omega) = 0 \mid \tilde{y}(\omega) = 0)$ | $Q_{2N}^{\%prd}$ |
| Residues correctly predicted | | $Q_2$ |
| **Per-Segment Metrics** | | |
| All observed TMH which are correctly predicted | | $Q_{tmh}^{\%obs}$ |
| All predicted TMH which are correctly predicted | | $Q_{tmh}^{\%prd}$ |
| Proteins for which all TMH are correctly predicted | | $Q_{ok}$ |

## 4.3 Results and Analysis

### 4.3.1 Transmembrane Helix Prediction

The results of the CRF model, on the test set, in predicting the transmembrane helix (TMH) regions are shown in Fig 7. These results were obtained from SBS [12]. The six different feature combinations that were used are shown in Table 9. A detailed description of the feature sets used was given in Section 3.

As we can see from the result in Fig 7, there are three main kinds of features which contribute the most in capturing useful information from the training data: Hydrophobic-based features, neighbouring-acids-based features and sequence-state-based features.

After adding single and double side neighbouring features, $Q_{ok}$ increases from 27% to 63% and $Q_2$ increases from 65% to 80%. After adding single and double shuffle neighbouring feature, $Q_{ok}$ increases to 75%. After adding hydrophobic neighbouring features, $Q_{ok}$ increases from 75% to 80% and $Q_2$ increases from 80% to 83%. Finally, after add sequence states features, $Q_{ok}$ increases up to 88% and $Q_2$ increases up to 84%, and these are the best results

Table 9: Enabled and disabled feature combination.

| Features | Exp 1 | Exp 2 | Exp 3 | Exp 4 | Exp 5 | Exp 6 | Exp 7 | Exp 8 |
|---|---|---|---|---|---|---|---|---|
| Basic | + | - | + | + | + | + | + | + |
| Properties | + | - | + | + | + | + | + | + |
| Hydrophobic Windows | - | - | + | + | + | + | + | + |
| Hydrophilic Windows | - | - | + | + | + | + | + | + |
| Single | - | +2 | +5 | +3 | +5 | +5 | +5 | +5 |
| Double | - | +1 | +1 | +1 | +3 | +3 | +3 | +3 |
| Single Shuffled | - | - | - | +3 | +6 | +6 | +6 | +6 |
| Double Shuffled | - | - | - | +1 | +3 | +3 | +3 | +3 |
| Single Hydrophobic | - | - | - | - | - | +3 | +6 | +6 |
| Double Hydrophobic | - | - | - | - | - | +1 | +3 | +3 |
| Single Hydrophilic | - | - | - | - | - | +3 | +6 | +6 |
| Double Hydrophilic | - | - | - | - | - | +1 | +3 | +3 |
| Border | - | - | - | - | - | - | + | + |
| Short Loops | - | - | - | - | - | - | + | + |
| Electronic | - | - | - | - | - | - | + | + |
| Groups | - | - | - | - | - | - | - | + |
| States | - | - | - | - | - | - | - | + |

*Basic* = Basic Amino Acid Features, *Properties* = Amino Acid Property Features, *Hydrophobic Windows* = Hydrophobic Window Features, *Hydrophilic Windows* = Hydrophilic Window Features, *Single* = Single Side Neighboring Amino Acid Features (with 1 to 5 neighbors from left or right), *Double* = Double Side Neighboring Amino Acid Features (with 1 to 3 neighbors from both sides), *Single Shuffled* = Single Side Shuffled Neighboring Amino Acid Features (with 1 to 6 neighbors from left or right), *Double Shuffled* = Double Side Shuffled Neighboring Amino Acid Features (with 1 to 3 neighbors from both sides). *Single Hydrophobic* = Single Side Hydrophobic Neighbouring Amino Acid Features (with 1 to 6 neighbors from left or right), *Double Hydrophobic* = Double Side Hydrophobic Neighbouring Amino Acid Features (with 1 to 3 neighbors from both sides). *Single Hydrophilic* = Single Side Hydrophilic Neighbouring Amino Acid Features (with 1 to 6 neighbors from left or right), *Double Hydrophilic* = Double Side Hydrophilic Neighbouring Amino Acid Features (with 1 to 3 neighbors from both sides). *Border* = Border Features, *Short Loops* = Short Loop Features, *Electronic* = Electron Transport Chain Features, *Groups* = Chemical Group Features, *States* = Sequence States Features.

we have attained.

This suggests three rules of the micro-level behaviour of proteins:

1. Hydrophobic values of amino acids is a significant factor of locating helical segments [31];

2. The helical structure of amino acids is greatly influenced by its neighbours;

3. Differences between the amino acids distributions in the various structural parts are one of the driving forces in the formation of the transmembrane helices [24, 13].

Experiment 8 (henceforth referred as “the CRFs model”) outperforms the other experiments in most categories and was selected for comparison with other prediction models.

We submitted the result of the CRF model on the test data set to the SBS and obtained a comparative ranking against other available methods. This is shown in Table 10, and indicates that the CRF model performed the best both on per-segment and per-residue accuracy.

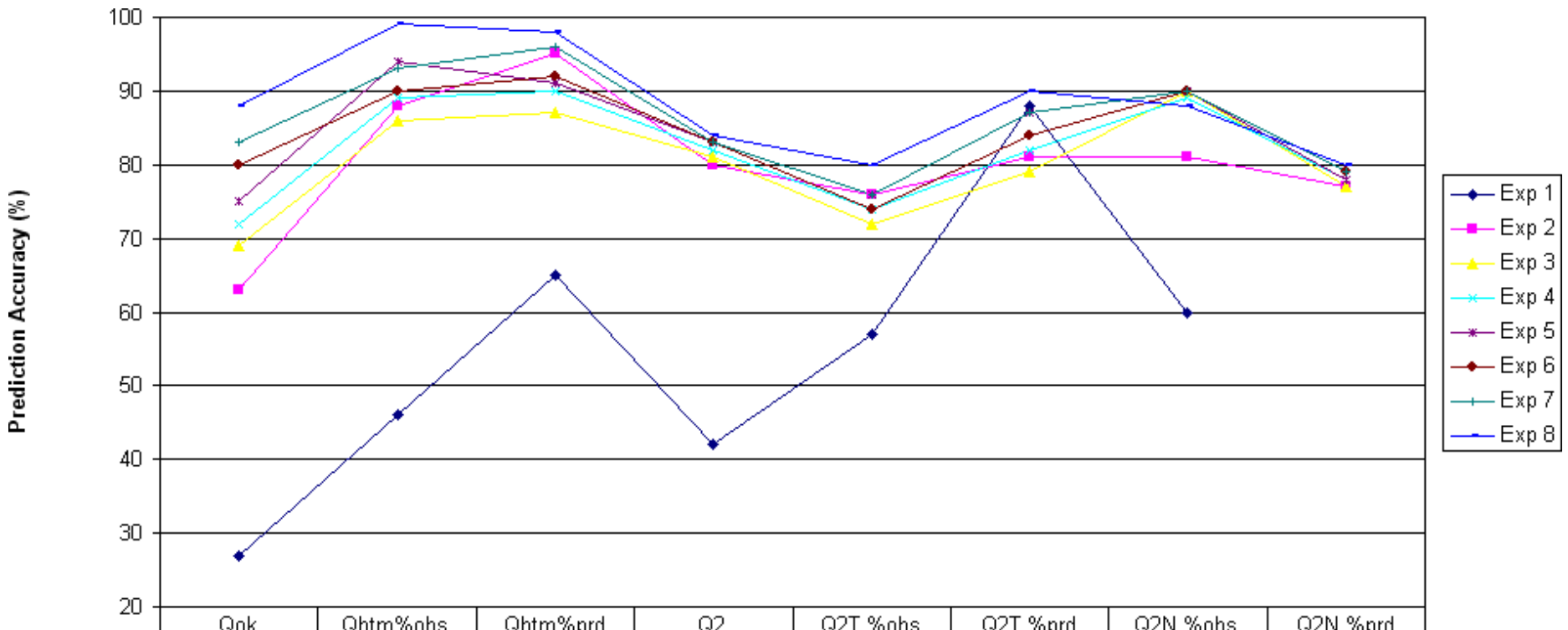

| | Qok | Qhtm%obs | Qhtm%prd | Q2 | Q2T %obs | Q2T %prd | Q2N %obs | Q2N %prd |
|---|---|---|---|---|---|---|---|---|
| Exp 1 | 27 | 46 | 65 | 42 | 57 | 88 | 60 | |
| Exp 2 | 63 | 88 | 95 | 80 | 76 | 81 | 81 | 77 |
| Exp 3 | 69 | 86 | 87 | 81 | 72 | 79 | 90 | 77 |
| Exp 4 | 72 | 89 | 90 | 82 | 74 | 82 | 89 | 78 |
| Exp 5 | 75 | 94 | 91 | 83 | 76 | 87 | 90 | 78 |
| Exp 6 | 80 | 90 | 92 | 83 | 74 | 84 | 90 | 79 |
| Exp 7 | 83 | 93 | 96 | 83 | 76 | 87 | 90 | 79 |
| Exp 8 | 88 | 99 | 98 | 84 | 80 | 90 | 88 | 80 |

Figure 7: Accuracy of the eight Experiments based on Feature sets shown in Table 9

#### 4.3.2 The Effect of Local Residue Distribution on Transmembrane Regions

We are interested in testing the hypothesis that the location of the transmembrane regions is determined by the composition of the amino acids in the transmembrane regions rather than their ordering. In order to test this hypothesis, we have carried out two experiments with different combinations of features turned on as shown in Table 11.

In Experiment 1 we have used single and double side neighboring amino acid features and in Experiment 2 we have used single and double side shuffled neighboring amino acid features (see Section 3 for an explanation of these features).

The prediction results of the two experiments are shown in Table 12 which suggest that the transmembrane helix regions are determined by the amino acid composition in the region rather than any specific ordering of the amino acids.

#### 4.3.3 Cytochrome c oxidase Protein Analysis

The Cytochrome c oxidase protein has one of the most complex structures known to date with a total of 28 transmembrane helices [33]. A detailed wet lab analysis of this complex, in terms of residue distribution and amino acid properties has been reported by Wallin et al. [33] (henceforth referred as the observed results). We have replicated the results using the CRF model with the features defined

Table 10: Prediction score comparison between 29 methods.

| Methods | Per-Segment $Q_{ok}$ | $Q_{tmh}^{\%obs}$ | $Q_{tmh}^{\%prd}$ | Per-Residue $Q_2$ | $Q_2^{\%obs}$ | $Q_{2T}^{\%prd}$ | $Q_2^{\%obs}$ | $Q_{2N}^{\%prd}$ |
|---|---|---|---|---|---|---|---|---|
| **CRFs** | **88** | **99** | **98** | **84** | **80** | **90** | **88** | **80** |
| PHDpsihtm08 | 84 | 99 | 98 | 80 | 76 | 83 | 86 | 80 |
| HMMTOP2 | 83 | 99 | 99 | 80 | 69 | 89 | 88 | 71 |
| DAS | 79 | 99 | 96 | 72 | 48 | 94 | 96 | 62 |
| TopPred2 | 75 | 90 | 90 | 77 | 64 | 83 | 90 | 69 |
| TMHMM1 | 71 | 90 | 90 | 80 | 68 | 81 | 89 | 72 |
| SOSUI | 71 | 88 | 86 | 75 | 66 | 74 | 80 | 69 |
| PHDhtm07 | 69 | 83 | 81 | 78 | 76 | 82 | 84 | 79 |
| KD | 65 | 94 | 89 | 67 | 79 | 66 | 52 | 67 |
| PHDhtm08 | 64 | 77 | 76 | 78 | 76 | 82 | 84 | 79 |
| GES | 64 | 97 | 90 | 71 | 74 | 72 | 66 | 69 |
| PRED-TMR | 61 | 84 | 90 | 76 | 58 | 85 | 94 | 66 |
| Ben-Tal | 60 | 79 | 89 | 72 | 53 | 80 | 95 | 63 |
| Eisenberg | 58 | 95 | 89 | 69 | 77 | 68 | 57 | 68 |
| Hopp-Woods | 56 | 93 | 86 | 62 | 80 | 61 | 43 | 67 |
| WW | 54 | 95 | 91 | 71 | 71 | 72 | 67 | 67 |
| Roseman | 52 | 94 | 83 | 58 | 83 | 58 | 34 | 66 |
| Av-Cid | 52 | 93 | 83 | 60 | 83 | 58 | 39 | 72 |
| Levitt | 48 | 91 | 84 | 59 | 80 | 58 | 38 | 67 |
| A-Cid | 47 | 95 | 83 | 58 | 80 | 56 | 37 | 66 |
| Heijne | 45 | 93 | 82 | 61 | 85 | 58 | 34 | 64 |
| Bull-Breese | 45 | 92 | 82 | 55 | 85 | 55 | 27 | 66 |
| Sweet | 43 | 90 | 83 | 63 | 83 | 60 | 43 | 69 |
| Radzicka | 40 | 93 | 79 | 56 | 85 | 55 | 26 | 63 |
| Nakashima | 39 | 88 | 83 | 60 | 84 | 58 | 36 | 63 |
| Fauchere | 36 | 92 | 80 | 56 | 84 | 56 | 31 | 65 |
| Lawson | 33 | 86 | 79 | 55 | 84 | 54 | 27 | 63 |
| EM | 31 | 92 | 77 | 57 | 85 | 55 | 28 | 64 |
| Wolfenden | 28 | 43 | 62 | 62 | 28 | 56 | 97 | 56 |

Obtained from the "Static benchmarking of membrane helix predictions" server (sorted by $Q_{ok}$)[12].

Table 11: The two different feature combinations used to test the effect of local residue distribution on TMH prediction.

| Exp#. | Basic | Properties | Single | Double | Border | Active Features |
|---|---|---|---|---|---|---|
| 1 | + | + | +5 | +3 | - | 803260 |
| 2 | + | + | - | - | - | 88594 |

in Experiment 8 (see Table 9).

We begin by predicting the distribution of individual residue type in the central membrane domain $\pm 10\AA$. The predicted and the observed results are shown in Fig 8(a) and indicate a high similarity between the observed and the predicted frequencies of the amino acids in the central membrane of the TMH regions.

Wallin et al. [33] also analyzed the frequency of the residues around the transmembrane helices based on the division of the amino acids into three groups: hydrophobic, polar and charged. We carried out a similar analysis using the CRF model and the results are shown in Fig 8(b). The concentration of hydrophobic residues around the central helix ($\pm$13 residues around 0) is in the frequency range between 20 to 35% (average of 24%). It is interesting to note that the frequency distribution of the polar residues is a virtual mirror

Table 12: Transmembrane prediction results using two different feature combinations.

| Per-Residue | | | | | | Per-Segment | | |
|---|---|---|---|---|---|---|---|---|
| Methods | $Q_2$ | $Q_{2T}^{\%obs}$ | $Q_{2T}^{\%prd}$ | $Q_{2N}^{\%obs}$ | $Q_{2N}^{\%prd}$ | $Q_{ok}$ | $Q_{tmh}^{\%obs}$ | $Q_{tmh}^{\%prd}$ |
| 1 | 83 | 72 | 84 | 93 | 78 | 75 | 87 | 91 |
| 2 | 83 | 75 | 86 | 89 | 77 | 75 | 92 | 83 |

Taken from the "Static benchmarking of membrane helix predictions" application written by Kernytsky et al. from University of Columbia, after submitting the prediction results of both experiments (sorted by $Q_2$ )[12].

image of the hydrophobic frequency along the central helix (average of 11.5%). The charged residues tend to appear with an average frequency of 6% outside the helix membrane and 3.5% inside. These results are compatible with physical experiments which show that highly polar and charged amino acids are energetically unfavorable inside the membrane, with the hydrophobic core of the cell membrane filling between the polar head groups. Recent studies by Hessa et al. [10] using a biological assay of membrane segments insertion suggest that not only the degree of hydrophobicity is important in structuring transmembrane helices but also the energetic stability of the helices, defined by the sum of individual contribution from each amino acid. The authors have demonstrated that the contribution to the total apparent free energy depends strongly on the position of each residue within the helix [10]. Generally the more the polar and charged amino acids are close to the membrane center, the larger the energetic penalty generally is. The accepted explanation is that electrostatic forces close to the polar head group and beyond the membrane help to stabilize the polar and charged groups in amino acids.

In support of this conjecture, we predicted the tendency of arginine (both polar and charged residue) around the transmembrane helix. We found that the tendency of arginine favorable to be located outside the membrane as demonstrated in Fig 8(c).

In contrast, we predicted the frequency around the membrane helix of isoleucine as an example of hydrophobic amino acids which according to observed experiments are favorable to appear in the center of the membrane. The results are demonstrated in Fig 8(d).

### 4.3.4 Approach

On the Cytochrome c oxidase experiments, we collected the distribution of individual residue type in the central membrane domain assuming that the central is on average $\pm 9$ residues from the transmembrane border. In Wallin's work, he has calculated the distribution on a region of $\pm$ 10$\mathring{A}$ [33], which we calculated using SwissPdbView [9], and found that on average the TMH region equals to 9 residues long. Wallin also referred his distribution figures to three different profiles: buried, intermediate and fully exposed residues. We compared our prediction results to the average values of the three profiles, since the sequence input that we used does not distinguish between these profiles.

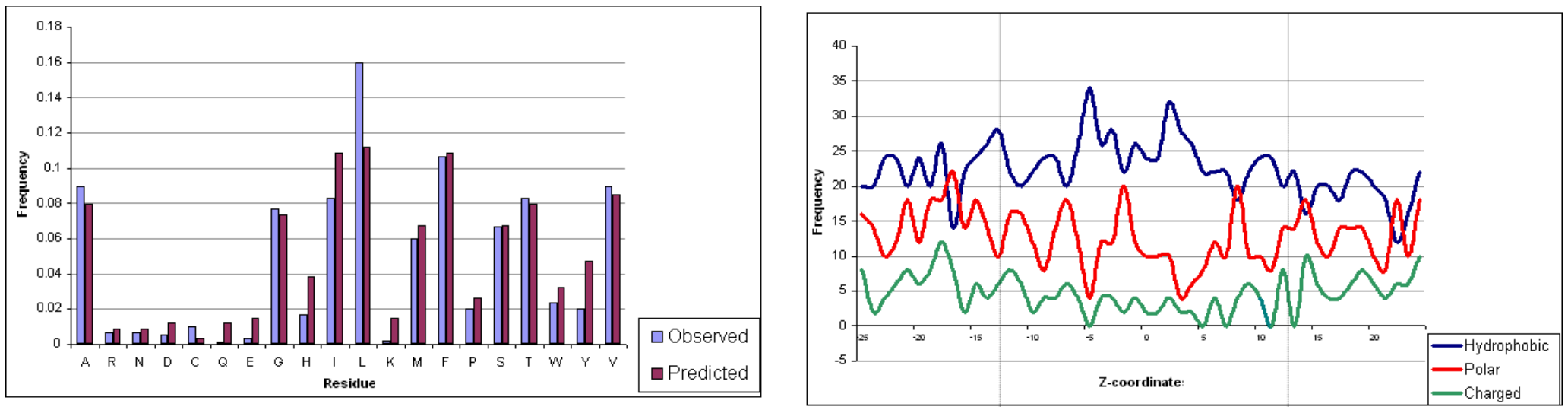

(a) Residue frequencies in the central membrane $\pm 10\mathring{A}$

(b) Distribution of residues types in central membrane region $\pm 25\mathring{A}$

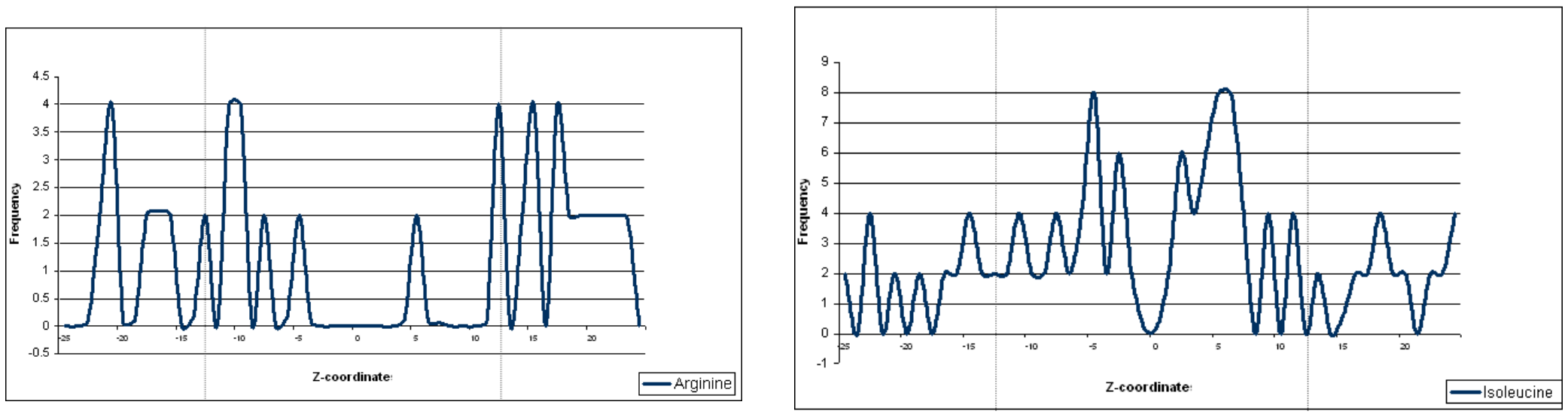

(c) Predicted Distribution of Arginine in TMH membranes $\pm$ 25 $\mathring{A}$ region.

(d) Predicted Distribution of Isoleucine in the TMH membranes

Figure 8: Cytochrome C Oxidase analysis. The results obtained from the CRF model and those reported by Wallin et. al [7], using "wet lab" experiments show a striking similarity. The distribution of Isoleucine and Arginine are consistent with the explanation by Hessa et. al[27]

The CRFs model was trained using data set consists of a set of benchmark sequences with experimentally confirmed transmembrane regions, which are significantly different, based on pairwise similarity clustering [20]. All occurrences of Cytochrome c oxidase sequences were removed from the training set to assure that the training set and test set are not overlapping.

# 5 Conclusions

Conditional Random Fields (CRF) are a class of statistical models which can be used to integrate disparate, overlapping and non-independent micro-level information to make predictions at the macro-level. In this paper we have applied the CRF model to predict the secondary structure of membrane proteins on the basis of their primary structure. Comparisons against twenty eight other models shown that the results of the CRF model are extremely accurate. We have also compared results of the CRF model against those obtained from wet lab experiments and have shown that we can replicate many of the important findings with striking accuracy. More importantly, the modular nature makes it possible for biologists to rapidly integrate and test new features (experiments), for relevance, in a mathematical model. The CRF model is extremely versatile and its use in bioinformatics is very likely to grow.

# 6 Appendix 1: CRF Definition and Model Derivation

Here we derive the CRF model in detail. A CRF model is an undirected markov model and the resulting probability distribution is equivalent to a conditional Gibbs distribution [16, 14].

Let $G = (V, E)$ be a graph, $Y = (Y_v)\ v \in V$ be a family of finite-valued random variables, and $X = (X_v)\ v \in V$ be a family of real-valued functions.

**Definition 1** *$(Y, X)$ is a conditional random field (CRF) if*

*1.* $P(Y|X) > 0 \ \forall \ Y$

*2. For all* $v \in V$

$$P(Y_v | Y_{V-v}, X) = P(Y_v | Y_{N(v)}, X)$$

*where* $N(v)$ *are the neighbors of* $v$ *in G. This is called the Markov property.*

**Definition 2** *Y is a Gibbs field if*

$$P(Y) = \frac{1}{Z}\exp\left(-\frac{1}{T}U(Y)\right)$$

*where Z is a normalizing constant, T is a constant called the temperature (which we assume is equal to one), and U is called the energy which can be expressed as:*

$$U(Y) = \sum_{c \in C} U_c(Y)$$

*where C is the set of all cliques (completely connected subgraphs) in G, and* $U_c$ *is the energy for a particular clique.*

According to the Hammersley-Clifford Theorem, a CRF is equivalent to a Gibbs field, thus

$$P(Y|X) \propto \exp\{U(Y,X)\} = \exp\{\sum_{c \in C} U_c(Y,X)\}^4 \qquad (5)$$

For sequential data, the graph $G$ is a chain and therefore the clique set $C$ consists of vertices ($C_1$) and edges ($C_2$), thus

$$U(Y,X) = \sum_{\{v\} \in C_1} V_1(Y_v, X) + \sum_{\{v,w\} \in C_2} V_2(Y_v, Y_w, X)$$

Using the notation of Lafferty [16] we can re-write Equation (5) as:

$$p_\theta(y|x) \propto exp\left(\sum_{e \in E,j} \lambda_j f_j(e, y|_e, x) + \sum_{v \in V,j} \mu_j g_j(v, y|_v, x)\right)$$

where $x$ is a data sequence, $y$ a label sequence, $y|_S$ is the set of components of $y$ associated with the vertices in subgraph $S$. The vectors $f$ and $g$ represent the local features with corresponding weight vectors $\lambda$ and $\mu$.
The joint distribution can be expressed in a slightly different form:

$$p_\theta(y|x) \propto exp\left(\sum_{i,j} \lambda_j f_j(y_{i-1}, y_i, x, i) + \sum_{i,k} \mu_k g_k(y_i, x, i)\right) \qquad (6)$$

$f_j(y_{i-1}, y_i, x, i)$ is a transition feature function of the entire observation sequence and the labels at positions $i$ and $i$-1 in the label sequence. $g_k(y_i, x, i)$ is a state feature function of the entire observation sequence and the label at position $i$ in the label sequence. $\theta = (\lambda_j, \mu_k)$ is estimated from the training data. We assume that the feature functions $f_k$ and $g_k$ are given and fixed [32]. The features serve as a "gateway" of incorporating biological information into the model.

[4] The minus sign has been absorbed in the function.

## 6.1 Feature Functions and Model Estimation

The feature functions constrain the conditional probability distribution $p(y|x)$. The satisfaction of a constraint increases the likelihood of the global configuration. Note that no feature independence assumption is made, and several dependent overlapping features are allowed. Different weights assigned to the parameters associated with the features, can be used to distinguish between important and irrelevant features.

An example of a feature is:

$$u(x, i) = \begin{cases} 1 & \text{if the amino acid at position } i \text{ is polar} \\ 0 & \text{otherwise} \end{cases}$$

An example of how the $u(x, i)$ feature is combined with possible state (label) information is as follows. If the current state in a state function, or the current and previous state in a transition function are satisfied then the feature function takes on the value $u(x, i)$ [32]. For example, the feature function $f_j(y_{i-1}, y_i, x, i)$ is assigned with the return value of the function $u(x, i)$, in case that the label at position $i-1$ and $i$ correspond to the $\alpha$-helix structure,

$$f_j(y_{i-1}, y_i, x, i) = \begin{cases} u(x, i) & \text{if } y_{i-1} = \alpha \text{ helix and } y_i = \alpha \text{ helix} \\ 0 & \text{otherwise} \end{cases}$$

In what follows, we generalize the transition functions to include state functions by writing:

$$g(y_i, x, i) = f(y_{i-1}, y_i, x, i)$$

We also define the sum of a feature over the sequence by:

$$f_j(x, y) = \sum_{i=1}^{n} f_j(y_{i-1}, y_i, x, i)$$

where $f_j(y_{i-1}, y_i, x, i)$ refers to either transition or state function [32]. Therefore, the probability of a label sequence $y$ given the observation sequence $x$ is of the form:

$$p_\lambda(y|x) = \frac{1}{Z_\lambda(x)} exp(\sum_j \lambda_j f_j(x, y)) \quad (7)$$

where

$$Z_\lambda(x) = \sum_y exp(\sum_j \lambda_j f_j(x, y)) \quad (8)$$

Equation 7 is the "closed-form" of the CRF model and it is important to note that it is normalized over the whole observation sequence $x$.

A function $\tilde{L}_{\tilde{p}}(\lambda)$ continues to remain concave because $L_{\tilde{p}}(\lambda)$ is concave and so is $-\sum_{j=1}^{K} \frac{\lambda_j^2}{2\sigma^2}$ and the sum of concave functions is concave. The $\sigma^2$ is a free parameter and large value of $\sigma^2$ corresponds to a higher penalty for larger $\lambda_j$ terms. However, as noted by Sutton et. al.[26] past research has shown that the final estimated $\lambda$ seems to be only moderately sensitive to the choice of $\sigma^2$.

## 6.2 Complete Algorithm: Parameter Estimation, Labeling and Complexity

The complete outline of the algorithm which calculates the maximum likelihood using the training data and then labels test sequences is shown in Algorithm 1. We walk through the steps and also note the computational complexity wherever relevant.

**Algorithm 1** Parameter Estimation and Labeling

```
1: Intialize
     – λ = < λ_j >_{j=1}^K   {:the feature parameters}
     – E                      {:convergence criterion for λ}
     – σ^2                    {:regularization parameter for ML}
2: Maximum Likelihood Estimation Phase:
3: For each active feature f_j, calculate  _{x,y} p̃(x, y)f_j(x, y) from training data
4: while convergence criterion E is not met do
5:    for all λ_k do
6:       Calculate  _{x,y} p̃(x)p_λ(y|x)f_j(x, y) using forward-backward algorithm
7:       Calculate  _{k=1}^K λ_k^2 / 2σ^2
8:    end for
9:    Combine Steps 3, 6, 7 (Eqn 9) and update λ using one iteration of the
      LBFGS Quasi Newton Method
10: end while
11: Labeling Phase on Test Data:
12: for all sequences in test data do
13:    Find the labeling y* = argmax_y P(y|x) for each sequence x using the
       Viterbi algorithm
14: end for
```

1. In Step 1, the parameters are initialized including $< \lambda_j >$, the convergence criterion parameter $E$ (which determines when the Quasi-Newton method will terminate) and the the regularization parameter $\sigma^2$.

2. In Step 4, the empirical expected value of each feature $f_j$ is calculated. This has to be done once for the training data. The cost of this step is $O(KNT)$, where $K$ is the number of parameter, $N$ is the number of training samples and $T$ is the average length of the training sequences.

3. In Step 6, the expected value of each $f_j$ under the model distribution is calculated for each training sample. This step requires the use of the forward backward algorithm [6, 26]. The total cost of this step (for all features) is $O(KNTL^2)$ where $K$, $N$ and $T$ are as before and $L$ is the number of labels (two in our case).

4. In Step 9, the $\lambda$ parameter is updated using the LBFGS Quasi Newton method. We have used an implementation LBFGS available under the Riso open source project and which uses the sparse matrix operations available from the Java COLT distribution [11].

5. In Step 13, each test sequence is labeled using the Viterbi algorithm. The cost of Viterbi algorithm on each sequence of length $T$ is is again $O(TL^2)$

Training for extremely large applications, which are common in computational linguistics can be quite prohibitive as noted both by Sutton [26] and Cohn [6]. However, in our case, we had over one and a half million features but less than two hundred proteins in the training set, training time was never more than five minutes per iteration on typical modern machine[5].

[5] Dell Latitude 410, 2GB RAM, 2GHz CPU